\documentclass[letterpaper, 10 pt, conference]{ieeeconf}  

\IEEEoverridecommandlockouts                              

\usepackage{amssymb}
\usepackage{graphicx}
\usepackage{graphics} 
\usepackage{epsfig} 
\usepackage{amsmath} 
\usepackage{mathtools}
\usepackage{algorithm}
\usepackage{algpseudocode}
\usepackage{comment}
\usepackage{dsfont}
\usepackage{array}
\usepackage{caption} 
\usepackage{multirow}
\usepackage{tabularx}

\newcommand{\etal}{\text{et al.}}
\newcolumntype{Y}{>{\centering\arraybackslash}X}

\title{\LARGE \bf
Anomaly-Aware Semantic Segmentation by Leveraging Synthetic-Unknown Data
}

\author{Guan-Rong Lu$^{1}$, Yueh-Cheng Liu$^{2}$, Tung-I Chen$^{1}$, Hung-Ting Su$^{1}$, Tsung-Han Wu$^{1}$, and Winston H. Hsu$^{1}$ \\ 
\thanks{$^{1}$National Taiwan University, $^{2}$National Tsing Hua University}
}

\begin{document}

\maketitle
\thispagestyle{empty}
\pagestyle{empty}

\begin{abstract}

Anomaly awareness is an essential capability for safety-critical applications such as autonomous driving. While recent progress of robotics and computer vision has enabled anomaly detection for image classification, anomaly detection on semantic segmentation is less explored. Conventional anomaly-aware systems assuming other existing classes as out-of-distribution (pseudo-unknown) classes for training a model will result in two drawbacks. (1) Unknown classes, which applications need to cope with, might not actually exist during training time. (2) Model performance would strongly rely on the class selection. Observing this, we propose a novel \textbf{Synthetic-Unknown Data Generation}, intending to tackle the \textbf{anomaly-aware semantic segmentation} task. We design a new \textbf{Masked Gradient Update (MGU)} module to generate auxiliary data along the boundary of in-distribution data points. In addition, we modify the traditional cross-entropy loss to emphasize the border data points. We reach the state-of-the-art performance on two anomaly segmentation datasets. Ablation studies also demonstrate the effectiveness of proposed modules. 

\end{abstract}

\section{INTRODUCTION}

The machine capability of realizing out-of-distribution (OOD) samples is essential for deploying learning systems to safety-critical real-world applications. 
For example, a self-driving vehicle is likely to encounter an object or a condition that is unknown or even does not exist during the training phase. 
However, modern deep learning-based methods, despite reaching remarkable performance on various computer vision and robotic tasks, are inclined to make overconfident predictions towards unknown samples and therefore result in myriad safety issues, as confirmed by recent studies \cite{amodei2016concrete,guo2017calibration,bendale2016towards,nguyen2015deep}.

Recent anomaly detection research has made progress in detecting OOD samples on image classification tasks~\cite{hendrycks2016baseline, liang2017enhancing,lee2017training,hendrycks2018deep,vyas2018out}. 
Auxiliary data, in particular, is proposed to actually enable the model capability of realizing unknown samples by fine-tuning a trained learning system, instead of manipulating predicted results (e.g., threshold-based methods~\cite{hendrycks2016baseline, liang2017enhancing}). 
In addition, prior works~\cite{lee2017training,hendrycks2018deep,vyas2018out} have demonstrated that training with auxiliary data sampled outside the training distribution can improve accuracy by effectively lowering the confidence scores of those anomalies on classification tasks. 
Nevertheless, various robotic applications such as auto-driving vehicles or medical robots require per-pixel prediction (i.e., semantic segmentation), which remains less explored, to make decisions. 
Moreover, it is inapplicable to directly adapt anomaly detection methods aiming for classification as they mostly assume one category per image, which is not the case for semantic segmentation tasks.

\begin{figure*}[t!]
    \centering
    \includegraphics[width=0.8\textwidth]{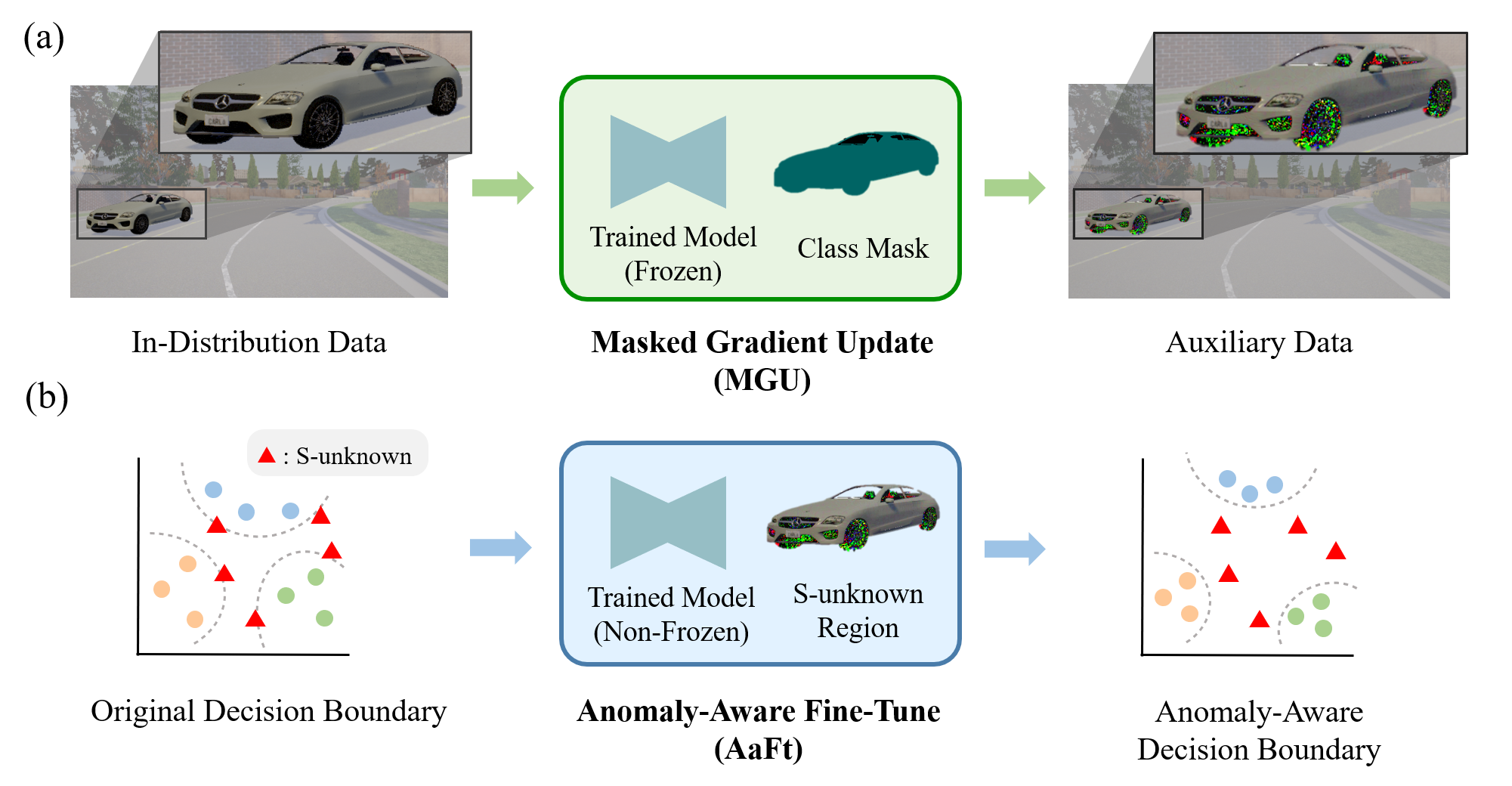}
    \caption{The two stages of our method. (a) shows the Masked Gradient Update method updates an in-distribution image to an auxiliary image that contains both synthetic-unknown regions and in-distribution regions. (b) shows that by finetuning with the \textit{boundary} synthetic-unknown regions, the decision boundaries become more precise to cluster the in-distribution data, which helps the model to distinguish anomalies. The red triangles represent the synthetic-unknown samples, and the points of three different colors present the samples of three different known classes.
    }
    \label{fig:fig1}
\end{figure*}

Several previous methods~\cite{hendrycks2018deep,li2020background} utilize existing classes that are out-of-distribution as auxiliary samples. 
Despite being able to tackle segmentation tasks theoretically, this principle of using ``known-unknown” classes assumes that they represent all potential anomalies and leads to two drawbacks. 
(1) It is impossible to train with all the unknown or even non-existing objects in the real world. 
(2) The performance of anomaly awareness strongly relies on the known-unknown class selection. 
For example, our experiment shows that the AUROC of models trained could vary from $80.7$ to $97.0$ with merely a different class selection of auxiliary samples (see Tab.~\ref{tab:pilot}).

Intuitively, out-of-distribution data covers the whole space excluding in-distribution data and it is infeasible to enumerate several classes with any selection method to overlay the remaining space.
Therefore, we hypothesize that the model could maximize the gain from auxiliary data points that are along the in-distribution boundaries. 
Hence, we propose a novel model-agnostic data generation algorithm to address the dilemma. 
As illustrated in Fig.~\ref{fig:fig1}(a), our Masked Gradient Update (MGU) adaptively generates \textit{synthetic-unknown} regions of auxiliary samples around the boundary of in-distribution data. 
In this way, the auxiliary data no longer suffer from the non-existing object issue and class selection bias. 
Furthermore, our MGU generated data could be utilized in any model or fine-tuning strategy. 
We design an anomaly-aware fine-tune (AaFt) module, as shown in Fig.~\ref{fig:fig1}(b), by fine-tuning on synthetic-unknown samples with a modified cross-entropy loss.
Following our principle of paying attention to the boundary, AaFt modifies the cross-entropy to emphasize the near boundary samples and further improves the performance. 
This experimental finding not only echos our hypothesis of the benefit from boundary samples but indicates that boundary-based approaches deserve further investigation.

The overall contributions of this work can be summarized as follows:
1) We are the first to quantify and tackle the selection bias of auxiliary data for anomaly segmentation.
2) The proposed method comprised of MGU and AaFt is model-agnostic and can train anomaly segmentation models in an adaptive manner.
3) We propose a novel entropy ratio loss to better separate the in- and out-of-distribution samples.
4) We achieve state-of-the-art results without building costly post-processing or ensemble models.


\section{Related work}
\label{sec:related_work}
\subsection{Anomaly Detection with Deep Neural Networks.}
%
Anomaly detection, also known as out-of-distribution (OOD) detection, is formulated as a problem to determine whether an input is in-distribution ($i.e.$, the training distribution) or out-of-distribution ($i.e.$, the other distributions).
The issue has been studied in several aspects, such as selective classification~\cite{geifman2017selective} and open-set recognition~\cite{bendale2016towards}.
Hendrycks \etal~\cite{hendrycks2016baseline} demonstrated that deep classifiers have a tendency to assign lower maximum softmax probability (MSP) to anomalous inputs than in-distribution samples.
Such property enables models to perform OOD classification by simple thresholding. 
Liang \etal~\cite{liang2017enhancing} further proposed temperature scaling and input preprocessing to make the MSP score become a more effective measurement for anomaly detection.
Monte Carlo dropout (MCDropout) has also been studied in previous works to measure network uncertainty~\cite{gal2016dropout, kendall2017uncertainties}, but the post-processing of such methods is rather expensive.
Some approaches based on ensemble learning have shown remarkable performance~\cite{lakshminarayanan2016simple}, but training multiple models requires considerable resources.
The threshold-based methods~\cite{hendrycks2016baseline} are notable for their computational efficiency, yet the pre-trained classifiers they used might fail to separate in- and out-of-distribution samples due to the overconfident prediction issue.
To address the overconfident prediction issue of the threshold-based methods~\cite{hendrycks2016baseline, liang2017enhancing}, \cite{lee2017training, hendrycks2018deep} trained anomaly detectors against an auxiliary dataset of outliers~\cite{goodfellow2014explaining} so the models could be capable of generalizing to unseen anomalies. 
In order to obtain more representative samples, \cite{lee2017training} leveraged GAN~\cite{mirza2014conditional} to synthesize adversarial examples around the in-distribution boundaries; \cite{li2020background} proposed a re-sampling strategy to construct a compact yet effective subset for training.
Additionally, \cite{hendrycks2018deep} proposed an extra loss term to penalize non-uniform predictions when out-of-distribution data are given, regularizing the learning so that OOD samples will have much lower confidence scores.
%

%
%
%
%
%

\subsection{Semantic Segmentation.}
%
Recent studies have made significant progress on semantic segmentation, which is one of the essential problems in computer vision and has shown great potential in a myriad of applications. 
FCN~\cite{long2015fully} is regarded as a fundamental work in semantic segmentation, which included only convolutional layers and can process images of arbitrary sizes.  
%
%
Feature pyramid~\cite{zhao2017pyramid} is also widely used in segmentation models to better capture multi-scale information.
DeepLabv3~\cite{chen2017rethinking} achieves admirable performance with computational efficiency by exploiting dilated convolutions.
Though these approaches have shown robustness and effectiveness, they have a tendency to make overconfident decisions as OOD inputs are given.
Such an insufficiency would seriously restrict these methods from deployment to real-world scenarios.

\subsection{Anomaly Segmentation.}
While prior attempts have studied OOD detection in multiple views, the task of anomaly segmentation remains rarely explored.
Wilddash dataset~\cite{zendel2018wilddash} was proposed to offer evaluations that reflect algorithmic robustness against visual hazards ($e.g.$, illumination changes, distortions, image noise), yet the OOD detection in \cite{zendel2018wilddash} still remains at the image-wise level.
Therefore, the CAOS benchmark~\cite{hendrycks2019benchmark} comprised of real-world driving scenes and synthetic anomalies is proposed, attempting to bridge the gap between anomaly detection and semantic segmentation.
%
%
Xia \etal~\cite{xia2020synthesize} proposed a framework consisting of both generative and comparison modules, identifying OOD regions by comparing the synthesized image to the original input.
Franchi \etal~\cite{franchi2020tradi} achieved state-of-the-art results by tracking the trajectory of weights and using an ensemble of networks.
Both \cite{xia2020synthesize, franchi2020tradi} demonstrated remarkable performance, however, these approaches require additional models to refine the output.
In this paper, we show our framework achieves superior performance by directly thresholding the softmax scores without exploiting generative modules, and such computational efficiency is highly desired in real-world applications.

\begin{figure*}[]
    \centering
    \includegraphics[width=0.8\textwidth]{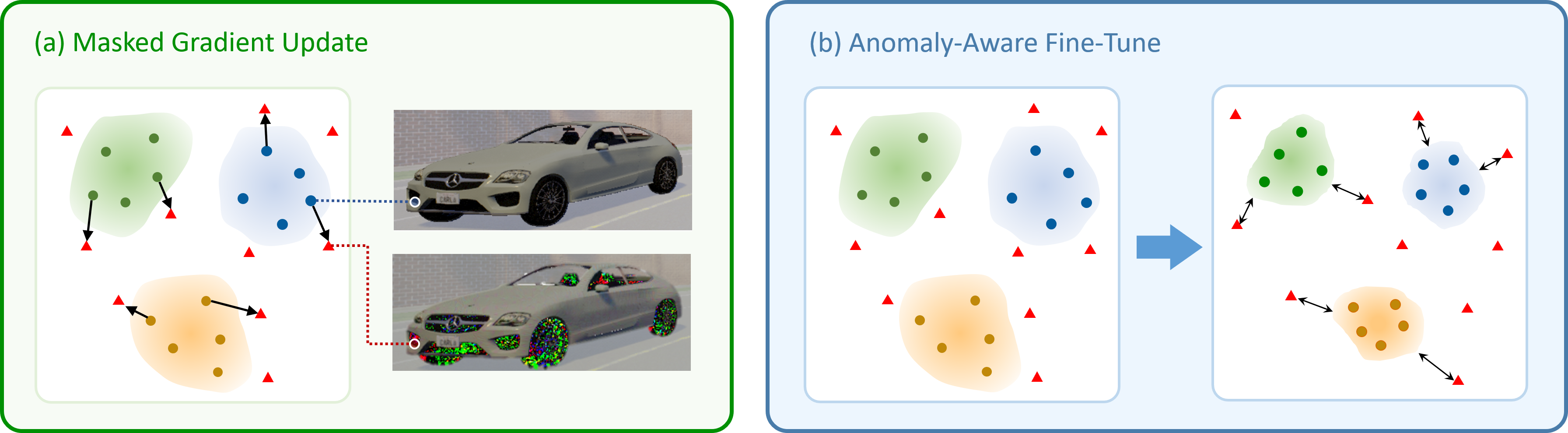}
    \caption{The concept of our method. In section (a), the in-distribution samples are modified to spread on the boundaries to generate boundary synthetic-unknown samples by Masked Gradient Updating. In section (b), by Anomaly-aware Fine-tuning with the additional unknown loss on synthetic-unknown regions, the model learns anomaly-aware decision boundaries. 
    }
    \label{fig:fig2}
\end{figure*}


\section{Method}
\label{sec:method}
%
In this paper, we aim to present an anomaly segmentation model based on existing semantic segmentation frameworks.
To tackle the Anomaly-aware Semantic Segmentation task, we propose two novel components. 
First, Masked Gradient Updating (\ref{sec:method:auxiliary}) generates auxiliary data on the boundary of in-distribution data without referencing additional existing data classes and labels. 
Second, Anomaly-aware Fine-tuning (\ref{sec:method:finetune}) learns anomaly-aware decision boundaries by adding additional loss for synthetic-unknown regions.

\subsection{Preliminary}
In this paper, we consider the problem of anomaly segmentation.
Unlike conventional OOD classification~\cite{hendrycks2016baseline, liang2017enhancing, lee2017training} treating a whole image as an anomalous sample, in anomaly segmentation, we must segment anomalous regions with pixel-level accuracy.
Let $D_k$ and $D_u$ denote two distinct distributions, \textit{known} and \textit{unknown}, respectively.
Those pixels of instances within training data are drawn from $D_k$, while the pixels of anomalous instances are sampled from $D_u$.
We further define a \textit{synthetic-unknown} distribution $D_s$ , which indicates the space of those synthetic pixels.
We assume a well-trained semantic segmentation model $f$ and the ground truth pixel labels $Z\in\mathbb{R}^{H\times W}$ associated with the input image $X\in\mathbb{R}^{C\times H\times W}$ are available.
For the convenience of interpretation, we define the well-known softmax probability as  
\begin{equation}
    S_i(X_{h, w}) = \frac{\exp(f_i(X_{h, w}))}{\begin{matrix}\sum^{N}_{j=1}\exp(f_j(X_{h, w}))\end{matrix}},
\end{equation}
where $f_i(X_{h, w})$ is the model output score of category $i$ at pixel $(h,w)$; $N$ is the number of classes in the dataset.
In the inference stage, we follow previous threshold-based methods~\cite{hendrycks2016baseline} to classify image pixels into anomalous objects $(D_u)$ if
\begin{equation}\label{eq:thresholding}
    \mathop{\max}_{i} S_i(X_{h, w}) \le \delta,
\end{equation}
where $\delta$ denotes the threshold.
An ideal anomaly segmentation model is expected to predict a class distribution $\{f_i(X_{h, w})|i=1,2...,N\}$ which is extremely close to the uniform distribution $\mathcal{U}(.)$ as an anomalous pixel $X_{h, w}\in D_u$ is given.
In the following sections, we will focus on explaining the training process of our anomaly segmentation framework.

\subsection{Generating auxiliary data by Masked Gradient Updating}\label{sec:method:auxiliary}
The auxiliary data plays a crucial role in our approach and is widely adopted in existing anomaly classification methods~\cite{lee2017training, hendrycks2018deep,li2020background}.
Nonetheless, the selection bias of known-unknown samples raises serious concern and remains rarely explored in anomaly segmentation (see Tab.~\ref{tab:pilot}).
Furthermore, within the more complicated problem setting of semantic segmentation, we cannot simply leverage a whole image drawn from another dataset as an auxiliary sample.

To solve the above problems, we propose Masked Gradient Update (MGU) addressing the aforementioned issues by producing auxiliary images comprised of both in- and out-of-distribution regions in an adaptive manner.
As shown in algorithm~\ref{alg:algorithm1}, we use $I_{\tilde{y}}$ to indicate the indices of those pixels belonging $\tilde{y}$, and $P_{\tilde{y}}$ denotes the indices of pixels predicted as $\tilde{y}$ correctly.
Given an initialized auxiliary image $X'$, we arbitrarily choose one of the classes existing in that image as the adversarial class $\tilde{y}\in\{1,2,...,N\}$.
Then, our proposed MGU will iteratively update the pixels included in $P_{\tilde{y}}$ conditioned on the cost function $L_{mgu}$.
The cost function is defined as
\begin{equation}
\mathcal L_{mgu} (f(X'), \tilde y) = \frac{1}{|P_{\tilde y}|}\sum_{(h, w) \in P_{\tilde y}}S_{\tilde y}(X'_{h, w}) .
\end{equation}
During the process, the parameters of $f$ are frozen and only the pixels within the target regions will be updated along the descending gradient direction of $L_{mgu}$.
By minimizing the softmax probability of $\tilde{y}$, those $X'_{h, w}\in I_{\tilde{y}}$ would be transformed into the synthetic-unknown pixels, which can also be regarded as the pseudo-unknown samples used for training an anomaly segmentation model.  
The auxiliary dataset can therefore be constructed by obtaining such synthetic-unknown images for each of the $N$ categories.
The concept of MGU is illustrated in Fig~\ref{fig:fig2}(a).
Unlike previous methods~\cite{hendrycks2018deep,li2020background} treating a known class as unknown objects, we aim to enforce the embeddings of synthetic-unknown instances to locate outside the decision boundary of known categories.
Punishing the pixels for having large $S_{\tilde y}(X'_{h, w})$ can be regarded as a process to enforce the embeddings of these pixels to move from their cluster center towards the decision boundary.
%
Note that once a pixel has been updated and predicted "incorrectly", it will immediately stop being updated.
%
Besides, the $\mathcal L_{mgu}$ encourages the pixels to be updated against their ground truth labels, so it can be expected the $S_{\tilde y}$ of each pixel to be small at the last few updates.
Therefore, we can expect the gradient of the last update of most pixels is too small to drag the samples far away from the boundary.
Such a property is highly desired because the synthetic-unknown instances would be the most discriminative pseudo-unknown samples that clearly identify the boundary of known categories.

\begin{algorithm}[t!]
\caption{Masked Gradient Update (MGU)}
\label{alg:algorithm1}
\begin{algorithmic}[1]
\Require
    semantic segmentation model $f$, input image $X$, adversarial class $\tilde{y}$, semantic segmentation label $Z$, set of image pixel indices $\Omega$, step size $\eta$
\State Initialize an auxiliary image $X' \gets X$
\State $I_{\tilde{y}}\gets\{(h, w) \;|\; (h, w)\in\Omega,\; Z_{h, w}\in\tilde{y}\}$ 
\State $P_{\tilde{y}}\gets I_{\tilde{y}}$
\While {$P_{\tilde{y}} \ne \emptyset$}
\State $P_{\tilde{y}}\gets\{(h, w) \;|\; (h, w)\in I_{\tilde{y}},\; \mathop{\arg\max}_{i} S_i(X'_{h, w})=\tilde{y}  \}$
        \For {$(h, w)$ in $P_{\tilde{y}}$}
            \State $X'_{h, w} \gets X'_{h, w} - \eta \nabla_{X'_{h, w}}\mathcal{L}_{mgu}(f(X'), \tilde{y})$
        \EndFor
\EndWhile

\end{algorithmic}
\end{algorithm}

\subsection{Anomaly-aware Fine-tuning}\label{sec:method:finetune}

After we generate the synthetic-unknown data through MGU, the next step is to fine-tune $f$, which is pre-trained on a normal semantic segmentation task, with those auxiliary data. 
As illustrated in Fig.~\ref{fig:fig2}(b), the idea is that we could learn to distinguish anomalies between in-distribution samples and the synthetic-unknown samples around the original model decision boundaries. 
Following the threshold-based method in Equation~\ref{eq:thresholding}, our goal is to lower the maximum probability value for those regions in $D_s$ while maintaining the original semantic segmentation output for regions in $D_k$.

For pixels in $D_s$, we apply the $\mathcal L_{KL}$ loss \cite{lee2017training,hendrycks2018deep,li2020background} to minimize KL divergence between the output probability distribution and uniform distribution. $\mathcal L_{KL}$ is defined as:
\begin{equation}
\begin{aligned}
\mathcal {L}_{KL}(f(X'_{h,w})) = -\sum\limits_{i=1}^N z \log \frac{S_i(X'_{h,w})}{z}
\end{aligned}
\end{equation}
where $X'_{h,w}$ is a pixel sampled from $D_s$, and $z=1/N$. This forces the output probability distribution towards uniform distribution for the anomaly areas.

Combined with the standard semantic segmentation cross-entropy loss $\mathcal L_k$, the objective function for fine-tuning is defined as:
\begin{equation} \label{eq:objective}
\begin{aligned}
\resizebox{0.43\textwidth}{!}{$\mathbb {E}_{X'_{h,w} \sim D_k}[\mathcal L_k(f(X'_{h,w}), Z_{h,w})] + \alpha \mathbb {E}_{X'_{h,w} \sim D_s}[\mathcal L_{KL}(f(X'_{h,w}))]$,}
\end{aligned}
\end{equation}
where $\alpha$ is weighting the hyper-parameter between the loss of standard semantic segmentation and anomaly segmentation. 
$\mathcal L_{k}$ calculates on all the pixels in $\mathcal D_k$, and $\mathcal L_{KL}$ calculates on all the pixels in the $\mathcal D_s$.

In this paper, we also explore another loss function option, $\mathcal L_{ER}$, for anomaly segmentation by measuring the entropy of softmax output. 
The entropy ratio loss $\mathcal L_{ER}$ is:
\begin{equation}\label{eq:er}
\begin{aligned}
\resizebox{0.43\textwidth}{!}{$\mathcal {L}_{ER}(f(X'_{h,w})) = \frac{-\sum\limits_{i=1}^{N} z\log(z)+r}{-\sum\limits_{i=1}^{N} \mathcal S_i(X'_{h,w})\log(S_i(X'_{h,w}))+r} - 1$,}
\end{aligned}
\end{equation}
where $r$ is a regularization term. 
The $\mathcal L_{ER}$ aims to maximize the output entropy of the pixels belonging to $D_s$.

Through the anomaly-aware fine-tuning, using the data generated from MGU, the model is able to perform the original semantic segmentation task and detect anomaly regions through thresholding.


\section{Experimental Results}
\label{sec:result}
\subsection{Training Setting}
In order to verify the effectiveness of our proposed method, we conduct experiments on two anomaly segmentation datasets (StreetHazards and BDD-Anomaly datasets).
StreetHazards dataset~\cite{hendrycks2019benchmark} is a large-scale anomaly segmentation dataset containing 5125 training images, 1031 validation images, and 1500 testing images with 12 object classes for self-driving applications.
Noted that for the testing images, a total of 250 unique anomalous objects appear in the image.
BDD-Anomaly dataset derives from the BDD100K semantic segmentation dataset~\cite{yu2018bdd100k}, a semantic segmentation dataset with diverse driving conditions.
We follow the experimental setup in~\cite{hendrycks2019benchmark}, regarding the classes \textit{motorcycle} and \textit{train} as anomalous objects.
The dataset contains 6688 training images, 951 validation images, and 361 testing images.
For network architectures, we use two state-of-the-art backbones, Deeplabv3~\cite{chen2017rethinking} and PSPNet~\cite{zhao2017pyramid}, to validate the performance of our method on the StreetHazards dataset, and we train a Deeplabv3 for anomaly-aware semantic segmentation on the BDD-Anomaly dataset.

For both backbones, we first train the model with Adam and the learning rate $0.0001$ for $1000$ epochs with batch size $256$ on standard semantic segmentation task.
Then, for auxiliary data generating, we choose $300$ images for each in-distribution class to apply MGU. 
The Deeplabv3 model is fine-tuned with Adam and the learning rate $0.0001$ for $10$ epochs with batch size $256$.
The PSPNet model is finetuned with Adam and the learning rate $0.0001$ for $30$ epochs with batch size $256$. 
The hyper-parameter $\alpha$ in Equation.~\ref{eq:objective} is $0.05$, and the regularization term $r$ in Equation.~\ref{eq:er} is $(-\sum_{i=1}^{12} z\log z) \times 0.01$, where $z$ is $1/12$ in this case for the both backbones.

\subsection{Evaluation Metrics}
We use the following three standard metrics to evaluate the effectiveness of the anomaly-aware semantic segmentation model. In this paper, we treat the unknown pixels as positive samples for all three metrics.

\textbf{AUPR} is the Area Under the Precision-Recall curve. In safety-critical applications, it is necessary to detect all anomalous objects. AUPR score focuses on evaluating the model performance on positive samples (i.e.\  anomalous samples), and it is more sensitive to positive and negative class imbalances. When positive samples are relatively small, AUPR is a more comprehensive indicator that reflects the data imbalance in the score. It is a threshold-independent metric, but the random guess score is the ratio between the number of positive and negative samples.

\textbf{AUROC} is the Area Under the Receiver Operating Characteristic curve. It is the probability for a positive sample to get a higher score than a negative sample. AUROC is also a threshold-independent metric, and the score of random guesses is always $50\%$ regardless of the ratio of the number of positive samples to the number of negative samples.

\textbf{FPR95} means the false positive rate at $95\%$ true positive rate. FPR at $x\%$ TPR tells how many false positives are inevitable for a given method to obtain the desired recall $x\%$.

\begin{table}[]
\centering
\setlength\extrarowheight{2pt}
\begin{tabular}{c|cccc}
\hline
 sub-experiment &
  \begin{tabular}[c]{@{}c@{}}AUPR$\uparrow$ \end{tabular} &
  \begin{tabular}[c]{@{}c@{}}AUROC$\uparrow$  \end{tabular} &
  \begin{tabular}[c]{@{}c@{}}FPR95$\downarrow$  \end{tabular} &
  \begin{tabular}[c]{@{}c@{}}AUPR\\ Random Guess \end{tabular} \\ \hline \hline
sub-exp 0       & \textbf{44.1} & \textbf{97.0} & \textbf{13.3} & 1.9    \\ 
sub-exp 1       & \textbf{7.5} & \textbf{80.7} & \textbf{79.1} & 1.6     \\ 
sub-exp 2       & 26.3 & 84.9 & 70.2 & 5.3    \\ 
sub-exp 3       & 29.4 & 92.9 & 30.5 & 1.5    \\ \hline
sub-average     & 26.8 & 88.9 & 48.3 & 2.6    \\ \hline
\end{tabular}
\caption{The result of the pilot study. The result shows that the three metric scores fluctuate among the four sub-experiment settings (different known-unknown classes selection).}
\label{tab:pilot}
\end{table}

\begin{figure*}[t!]
    \centering
    \includegraphics[width=\textwidth]{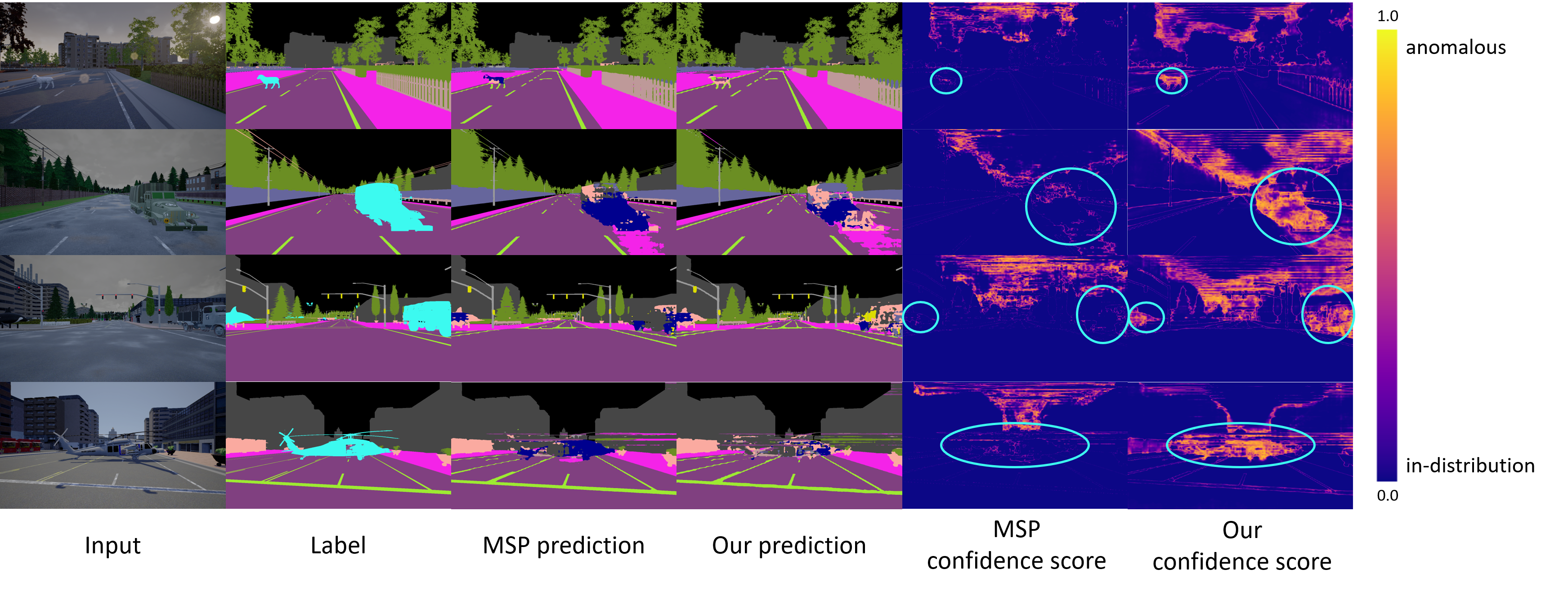}
    \caption{Visualizations on the StreetHazards Dataset. The sky blue regions in the labels are anomalous objects. The higher score (close to yellow) in the heat maps represents the higher confidence, which means the pixel belongs to anomalous objects. The model trained with our method can segment the anomalous objects, which is unable to detect in MSP.
    }
    \label{fig:fig3}
\end{figure*}

\subsection{Selection Bias}\label{sec:result:selection}
To investigate the selection bias of known-unknown-based methods~\cite{hendrycks2018deep,li2020background}, we conduct a pilot study with the same settings except for merely different instance categories serving as known-unknown samples.
Specifically, the $12$ classes in StreetHazards dataset~\cite{hendrycks2019benchmark} are split into $4$ subsets.
Each subset consists of $3$ classes without overlaps, and in each sub-experiment, one of the subsets will be chosen as the \textit{anomaly subset}.
Those instances in an image belonging to the categories of the anomaly subset should be regarded as anomalies for training.
Such an experimental setting is consistent with those methods using auxiliary data~\cite{hendrycks2018deep,li2020background}, except that we explore anomaly segmentation rather than classification.

As shown in Tab.~\ref{tab:pilot}, the significantly fluctuating performance indicates that known-unknown class selection heavily influences the results. The results reveal our hypothesis of \textit{known-unknown class selection correlates to model performance}. Therefore, known-unknown classes could barely represent the out-of-distribution regions. On the other hand, our proposed MDG can adaptively generate synthetic-unknown instances based on observing distribution instead of choosing particular categories as known-unknown instances.

\subsection{Result}
Tab.~\ref{tab:result} demonstrates the experimental result on the StreetHazard dataset. Using the backbone of both Deeplabv3 and PSPNet, our MGU+AaFt with entropy ratio loss reaches state-of-the-art performance on AUPR and AUROC metrics while having a competitive FPR95 score. 
As there are only $1.5\%$ unknown pixels in the StreetHazards dataset, metrics that do not consider data imbalance might underestimate the significance of detecting unknown pixels. The impressive AUPR scores ($5.3\%$ and $1.6\%$ enhancement with Deeplabv3 and PSPNet), which is the only one regarding data imbalance, suggest the capability of anomaly awareness of our proposed method. 
Besides, our method also performs well on the BDD-Anomaly dataset, as shown in Tab.~\ref{tab:bddresult}. The results reveal that our method outperforms the best previous score by $6.5\%$ in AUPR.
The results indicate the effectiveness and the robustness of our proposed components. 

Also, more than achieving remarkable performance, our method is more resource-efficient. Prior works either require multiple models (Deep Ensemble~\cite{lakshminarayanan2016simple}, TRADI~\cite{franchi2020tradi}) or reconstruct original testing images (SynthCP~\cite{xia2020synthesize}). Additionally, our approach is model-agnostic and has huge potential to further boost the performance of future models or applications.

Fig.~\ref{fig:fig3} demonstrates several cases on the StreetHazard dataset. MSP method alone is unable to tackle anomalies on the roads and leads to critical safety issues. With the same backbones, we finetune the model with our generated auxiliary data and enable the model to recognize unknown objects. 

\begin{table}[]
\centering
\setlength\extrarowheight{2pt}
\begin{tabular}{@{}p{0.23\textwidth}|ccc@{}}
\hline
Method & AUPR$\uparrow$ & AUROC$\uparrow$ & FPR95$\downarrow$ \\ \hline \hline
Dropout~\cite{gal2016dropout}           & 7.5 & 79.4   & 69.9  \\ 
MSP~\cite{hendrycks2016baseline} ($\mathcal L_k$ only)        & 6.6 & 87.7   & 33.7  \\ 
MSP+CRF~\cite{hendrycks2019benchmark}   & 6.5 & 88.1   & 29.9  \\ 
SynthCP~\cite{xia2020synthesize}        & 9.3 & 88.6   & 28.4  \\ 
TRADI~\cite{franchi2020tradi}   & 7.2 & 89.2   & \textbf{25.3}  \\ 
Deep Ensemble~\cite{lakshminarayanan2016simple}             & 7.2 & 90.0   & 25.4  \\ \hline
MGU+AaFt($\mathcal L_k + \mathcal L_{ER}$, pspnet)           & 10.9 & 89.5  & 32.9 \\
MGU+AaFt($\mathcal L_k + \mathcal L_{KL}$, deeplabv3)           & 13.2 & 90.4  & 34.0 \\
MGU+AaFt($\mathcal L_k + \mathcal L_{ER}$, deeplabv3)           & \textbf{14.6} & \textbf{92.2}  & 25.9 \\ \hline
\end{tabular}
\caption{The anomaly segmentation results of the StreetHazards dataset. We achieve state-of-the-art results in anomaly segmentation by giving big progress in AUPR score and AUROC score while having a competitive score in FPR95.}
\label{tab:result}
\end{table}

\begin{table}[]
\centering
\setlength\extrarowheight{2pt}
\begin{tabular}{@{}p{0.23\textwidth}|ccc@{}}
\hline
Method &
  \begin{tabular}[c]{@{}c@{}}AUPR$\uparrow$\end{tabular} &
  \begin{tabular}[c]{@{}c@{}}AUROC$\uparrow$\end{tabular} &
  \begin{tabular}[c]{@{}c@{}}FPR95$\downarrow$\end{tabular} \\ \hline \hline
Dropout~\cite{gal2016dropout}          & 6.5 & 83.7   & 33.4  \\ 
MSP~\cite{hendrycks2016baseline}        & 6.3 & 84.2   & 31.9  \\ 
MSP+CRF~\cite{hendrycks2019benchmark}   & 8.2 & 86.3   & 26  \\ 
TRADI~\cite{franchi2020tradi}   & 5.6 & 86.1   & 26.9  \\ 
Deep Ensemble~\cite{lakshminarayanan2016simple}             & 6 & 87   & \textbf{25}  \\ \hline
MGU+AaFt($\mathcal L_k + \mathcal L_{ER}$, deeplabv3)           & \textbf{14.7} & \textbf{91.5}  & 30.8 \\ \hline
\end{tabular}
\caption{The anomaly segmentation results of the BDD-Anomaly dataset. The experiment shows that our method outperforms the best previous score in AUPR by $6.5\%$.}
\label{tab:bddresult}
\end{table}

\subsection{Ablation Study}
In the ablation study, we measure the effectiveness of each designed module. As shown in Tab.~\ref{tab:result}, by training with our MGU generated, we boost the MSP (first block, second row) from 6.6 AUPR to 13.2 (second block, second row). Considering MSP is a relatively simple method, the performance gain is impressive. Additionally, it reveals possible further performance improvement of training modern models with MGU generated data.

Our entropy ratio loss $\mathcal L_{ER}$ aims to investigate the influence of loss usage, further improving the performance from 13.2 AUPR to 14.6 (second block, third row). As a step toward, the result suggests a potential research direction for future work.

\subsection{Threshold Analysis}
\begin{figure}[t!]
    \centering
    \includegraphics[width=0.45\textwidth]{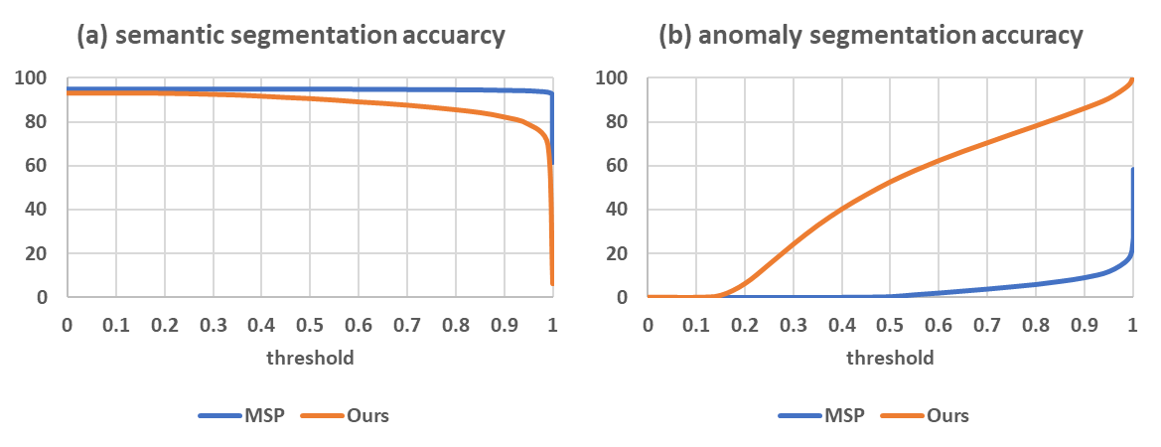}
    \caption{The comparison of semantic and anomaly segmentation accuracy. Our method not only improves anomaly segmentation accuracy but also maintains well semantic segmentation performance.
    }
    \label{fig:threshold}
\end{figure}

The AUPR, AUROC, and FPR95 metrics measure the relative output confidence scores between in-distribution pixels and anomalous pixels regardless of the correctness of semantic segmentation.
In this section, we introduce how thresholding influences the performance of semantic and anomaly segmentation.
Fig.~\ref{fig:threshold} shows the semantic and anomaly segmentation accuracy of our method and MSP. 
As increasing $\delta$, more pixels are classified as anomalies. 
Note that the MSP model has not been fine-tuned with anomaly data, so ideally adjusting the threshold $\delta$ affects little in the semantic segmentation performance; however, it will encounter the overconfidence issue when fed with anomalous inputs.
Overall, as increasing $\delta$, the anomaly-aware semantic segmentation model trained with our method can significantly improve the anomaly segmentation accuracy, while the model still maintains desirable semantic segmentation performance.

\section{Conclusion}
\label{sec:conclusion}
In this paper, we study a practical yet less explored anomaly-aware semantic segmentation task and, first-ever, leverage auxiliary data to tackle it. Observing the drawbacks of traditional known-unknown-based auxiliary data approaches, we propose a novel MGU module to generate samples along the border of in-distribution data. Moreover, we investigate the usage of loss function when fine-tuning and design an entropy ratio loss. Experimental results show the effectiveness of our proposed modules. We are optimistic that our work could pave a new path for future research.

\bibliographystyle{IEEEtran}
\bibliography{reference}

\end{document}